\documentclass[lettersize,journal]{IEEEtran}
\usepackage{amsmath,amsfonts}
\usepackage{algorithmic}
\usepackage{algorithm}
\usepackage{array}
\usepackage[caption=false,font=normalsize,labelfont=sf,textfont=sf]{subfig}
\usepackage{textcomp}
\usepackage{stfloats}
\usepackage{url}
\usepackage{verbatim}
\usepackage{multirow}
\usepackage{graphicx}
\usepackage{cite}
\usepackage{makecell}
\usepackage{array} 
\usepackage{threeparttable}
\usepackage{enumitem}
\usepackage{csquotes}
\usepackage[colorlinks,
linkcolor=red,
anchorcolor=blue,
citecolor=green
]{hyperref}

\begin{document}

\title{SeCG: Semantic-Enhanced 3D Visual Grounding via Cross-modal Graph Attention } 

\author{Feng Xiao, Hongbin Xu, Qiuxia Wu, Wenxiong Kang, ~\IEEEmembership{Member, ~IEEE} \thanks{Feng Xiao, Hongbin Xu, and Wenxiong Kang are with the School of Automation Science and Engineering, South China University of Technology, Guangzhou, China, 510006. Qiuxia Wu is with the School of Software Engineering, South China University of Technology, Guangzhou, China, 510006.
Wenxiong Kang is the corresponding author (email: auwxkang@scut.edu.cn).

The codes are released at \href{https://github.com/onmyoji-xiao/3dvg_SeCG}{{https://github.com/onmyoji-xiao/3dvg\_SeCG}}. }}


\maketitle

\begin{abstract}
3D visual grounding aims to automatically locate the 3D region of the specified object given the corresponding textual description. Existing works fail to distinguish similar objects especially when multiple referred objects are involved in the description. Experiments show that direct matching of language and visual modal has limited capacity to comprehend complex referential relationships in utterances. It is mainly due to the interference caused by redundant visual information in cross-modal alignment. To strengthen relation-orientated mapping between different modalities, we propose SeCG, a semantic-enhanced relational learning model based on a graph network with our designed memory graph attention layer. Our method replaces original language-independent encoding with cross-modal encoding in visual analysis. More text-related feature expressions are obtained through the guidance of global semantics and implicit relationships. Experimental results on ReferIt3D and ScanRefer benchmarks show that the proposed method outperforms the existing state-of-the-art methods, particularly improving the localization performance for the multi-relation challenges. 
\end{abstract}

\begin{IEEEkeywords}
3D visual grounding, visual-language learning, point cloud, graph attention, referential relationship
\end{IEEEkeywords}

\section{Introduction}

\begin{figure}[!t]
   \centering
    \includegraphics[width=1\linewidth]{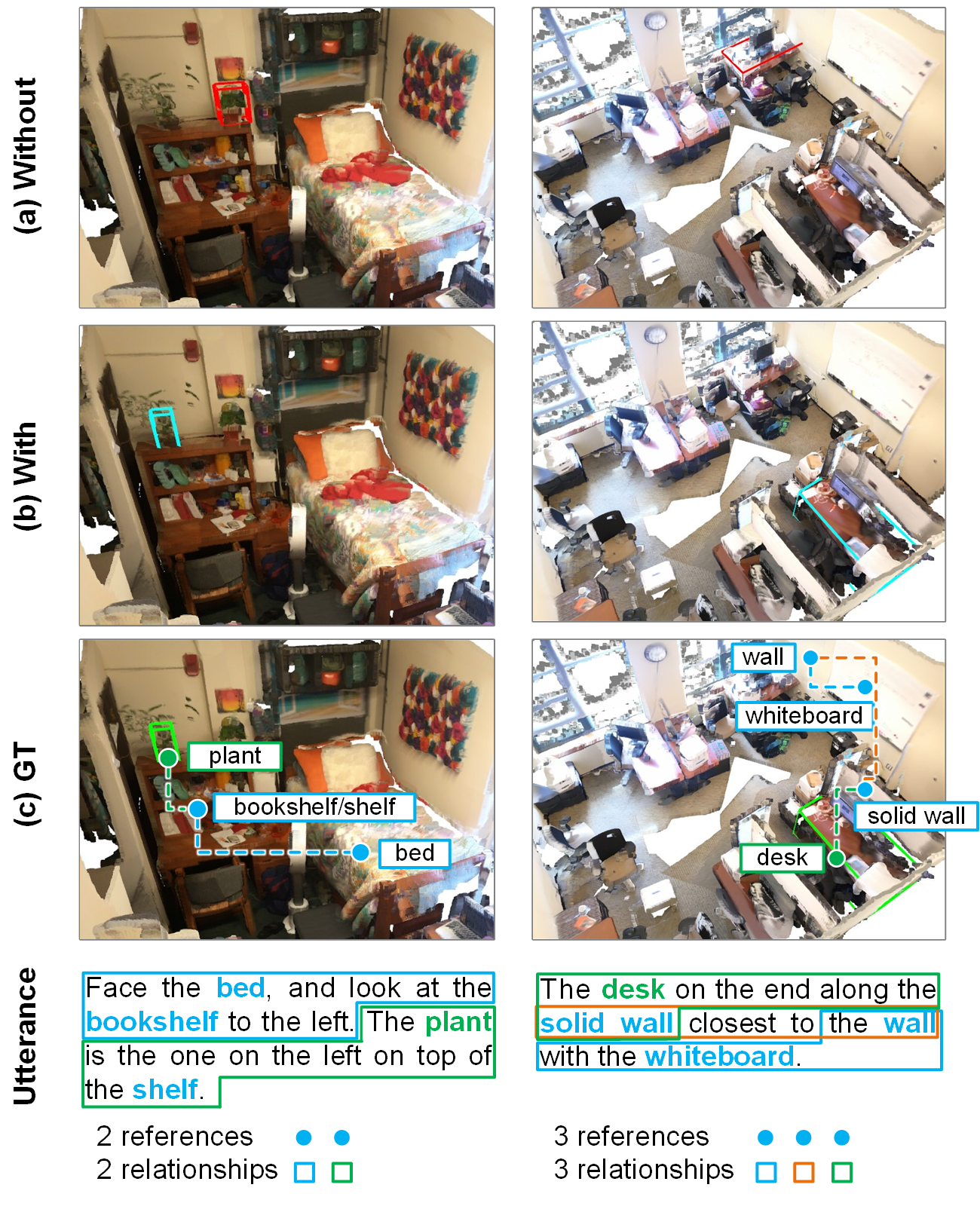}
    \caption{Comparison of results without(a) and with(b) our multi-relation improvement, (c) shows the ground truth and related objects. The green words in the utterances are target names and the blues are references. The decomposed pairwise relationships are framed on the text, corresponding to the dashed lines of the same color in above pictures.}
    \label{fig1}
    \vspace{-0.4cm}  
\end{figure}

\IEEEPARstart{V}{ision} and language are two critical modalities for computers to understand real 3D scenes and solve deep problems \cite{du2022survey}. A plethora of studies on modalities of interaction between vision and language provide the substantial prerequisite for various applications such as autonomous driving, robotics, and remote sensing \cite{lahoud20223d}. With the flourishing of multi-modal learning, 3D visual grounding is a novel challenging task aimed at empowering the computer to find the correspondence between human language and 3D visual information.

Nowadays the core of well-explored 3D visual grounding tasks is the perception of referential relationships. It means the target object is found by describing its position relative to the referred objects. In terms of single modality, 3D object detection and natural language processing have achieved great success in their respective domains. These existing models exactly support the recognition of fine-grained objects and the parsing of natural sentences \cite{guo2020deep,yin2022survey}. Consequently, the challenge of 3D visual grounding essentially depends on cross-modal alignment, where the uncertain viewpoints and interference from other similar objects increase the difficulty. This intelligent algorithm is required to not only distinguish the target and the referred objects at the semantic level but also find the unique object based on the complex orientation relationship in the sentence.

Existing models are prone to misselect the distractors with weak understanding of the referential dependence especially involved of multiple referred objects \cite{achlioptas2020referit3d,chen2020scanrefer,roh2022languagerefer,yang2021sat,bakr2022look,huang2022multi}. Directly matching two modal features of independent encoding in 3D visual grounding is deficient to understand complex relationships. As shown in Fig \ref{fig1}, each utterance contains multiple objects that are spatially related to the target directly or indirectly. In (a), the basic method of matching separately extracted text and visual features incorrectly locates the similar distractor, which is mainly due to insufficient perception of some referential relationships. It is worth noting that the definition of the target also depends on the multi-level relationships between other objects such as \enquote{bed} and \enquote{bookshelf}, \enquote{wall} and \enquote{whiteboard}. Take the second scene as an example, there are three objects involved in the definition of the target \enquote{desk}, and its specific position is determined by the progressive understanding of the three marked relationship phrases. In (b), the improved method pays more attention to the simultaneous perception of multiple references and locates correct targets.

To handle the aforementioned issue, we tend to improve the cross-modal alignment for descriptions with multiple referred objects from two perspectives: (i) \textbf{Relational learning}. Modeling the object relationships before visual-language matching is conducive to the subsequent integration of orientation clues. Adding language information in the autonomous learning of relationships can guide the visual encoding to a specific direction about salient objects, while visual encoding in most existing models is language-independent. (ii) \textbf{Semantic enhancement}. In human thought patterns, we often first localize the salient objects of the mentioned categories when facing a long description. That means before analyzing the inherent appearance properties and related connections, the semantic category completes the preliminary screening. The prior semantic knowledge can be utilized in early encoding to extract more associated features for relation-level perception. 

\begin{figure*}[t]
    \centering{\includegraphics[width=1\linewidth]{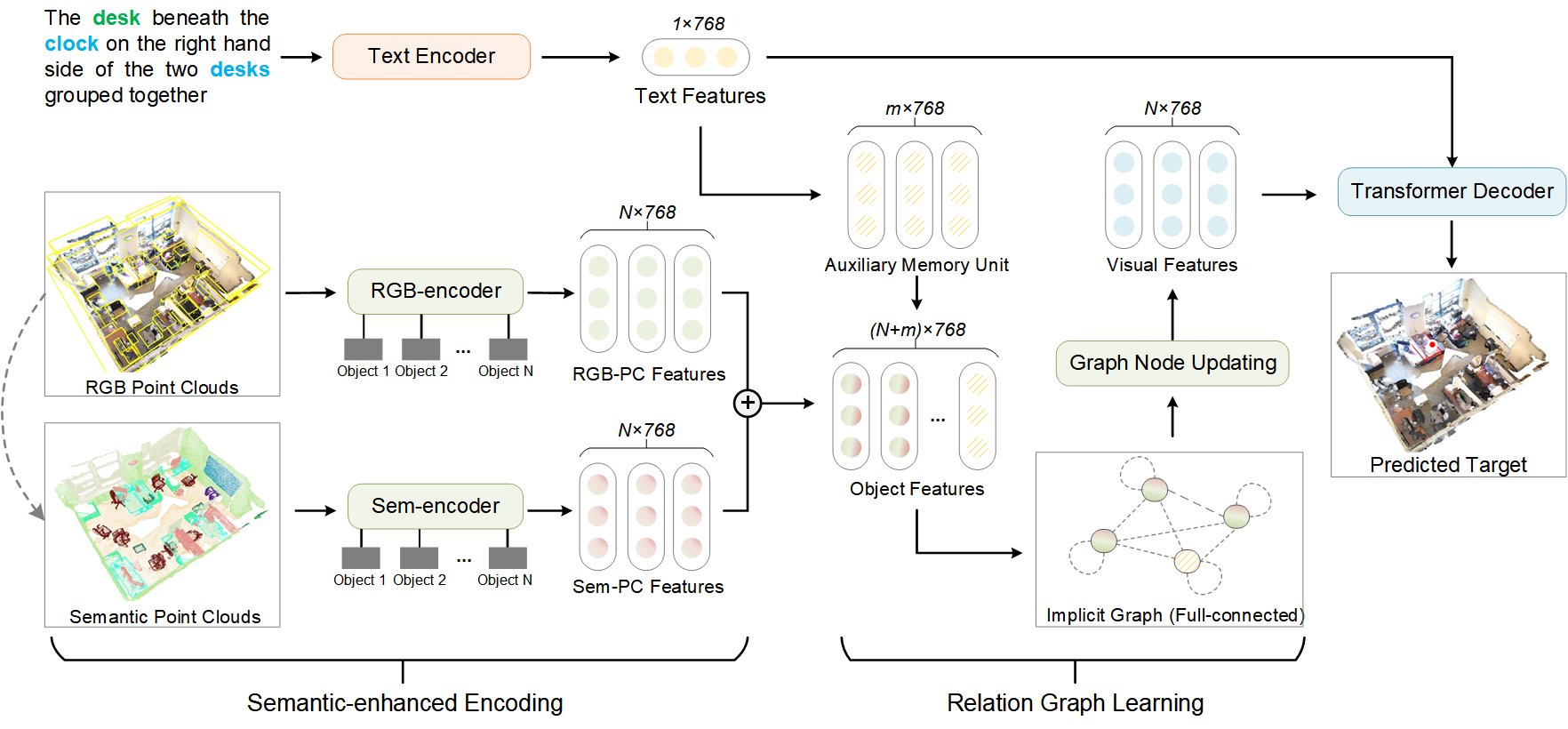}}
	\caption{\textbf{The overall architecture of our proposed model, SeCG}, consists of four modules: semantic-enhanced encoding, relation graph learning, text encoding, and Transformer decoding. The input data are segmented instance point clouds and referential utterance, the semantic point cloud is generated as an intermediate result. In each scene, $N$ objects are used to construct the relation graph. The text generates a $m$-dimensional memory matrix in each graph updating layer.}
   \vspace{-0.4cm}
    \label{fig2}
\end{figure*}

In this paper, we propose \textbf{SeCG}, a semantic-enhanced relational learning model based on graph attention for 3D visual grounding. The overall pipeline is shown in Fig \ref{fig2}. \textbf{For relational learning}, encoded objects from point clouds are constructed as nodes into a novel graph attention network(GAT) to learn implicit relationships. Our graph network is capable of extracting the attention features relevant to referential descriptions through a multi-modal memory unit. In order to facilitate the adaptability of the model to various viewpoints, we embed relative position encoding of multiple views into graph attention calculations. \textbf{For semantic enhancement}, each node in the graph is prompted to simultaneously aggregate deep features from two modes of expression, the original RGB point cloud, and the semantic point cloud. The semantic point cloud is a high-level expression without color and texture, which makes the encoder focus more on object position and category and provides direct guidance for cross-modal alignment on relational scene comprehension. Finally, the visual encoding features and language encoding features are fed into a transformer decoder to find the corresponding target. 

The main contributions are summarized as follows:
\begin{itemize}[leftmargin=*]
\item{We propose a semantic-enhanced visual grounding model with cross-modal graph attention, focusing on the challenging localization with multiple referred objects.}
\item{We design a novel graph attention layer for implicit relational learning, which introduces language-guided memory units and multi-view geometrical assistance.}
\item{We first exploit prior semantic knowledge in point cloud encoding by constructing a new expression, synchronously perceiving more directed information from referential languages.}
\item{Our method is tested on ReferIt3D \cite{achlioptas2020referit3d} and ScanRefer \cite{chen2020scanrefer}, outperforms the existing state-of-the-art methods.}
\end{itemize}

\section{Related Work}
\subsection{3D Visual Grounding}
Visual grounding is a multidisciplinary task integrating computer vision and natural language processing, has been extended to the 3D field in recent years. The mainstream reasoning pipelines are generally divided into two camps, two-stage methods where the objects are extracted from front detectors and one-stage methods adjusting the predicted regions synchronizing with multi-modal alignment \cite{huang2022deconfounded}. 3D visual grounding task is conducted on scene point clouds, devoted to matching the sole target from a lot of objects by the referential description. Unlike 2D images, point clouds are sparse and noisy, and lack dense texture and structured representation, seriously limiting the migration of outstanding 2D localization methods with pixel-level visual encoding \cite{deng2023transvg++,du2022visual,chen2022multi}. There are also some methods that jointly train visual grounding and captioning models to achieve complementary improvement \cite{cai20223djcg,chen2022d}.

ScanRefer \cite{chen2020scanrefer} is the first work to construct a large-scale 3D location dataset with free-form descriptions and develop a two-stage network architecture. Then ReferIt3D \cite{achlioptas2020referit3d} provides other two large-scale and complementary datasets, Nr3D and Sr3D, which only focus on scenes with multiple instances of the same fine-grained class. Based on the above baselines, many two-stage grounding methods have been proposed. SAT \cite{yang2021sat} utilizes semantic features of 2D projection from a pre-trained detector in training to learn a better 3D object representation, while LAR \cite{bakr2022look} directly synthesizes 2D images of each object as auxiliary semantics. TransRefer3D \cite{he2021transrefer3d} is the first to introduce the Transformer \cite{vaswani2017attention} architecture, establishing entity-and-relation aware attention to distinguish the referent corresponding to the same word. The research of LanguageRefer \cite{roh2022languagerefer} pays more attention to view-dependent utterances and verifies the positive significance of viewpoint correction. Also for the view rotation challenge, MVT \cite{huang2022multi} fuses multi-view position encoding with point cloud features and obviously improves the overall performance at a relatively low cost by rotating box coordinates. In contrast, ViewRefer \cite{guo2023viewrefer} rotates the point clouds to multiple views to extract visual features, and leverages a large-scale language model to expand the view-related text.

Different from the detection-then-matching pipelines, one-stage grounding models still predict or adjust bounding boxes in final multi-modal decoding. BUTD-DETR \cite{jain2022bottom} outputs the target box by an extra prediction head of the decoder, which means text understanding can indicate detection results. EDA \cite{wu2023eda} similarly accomplishes the box prediction in the decoding stage, preventing imbalance and ambiguity learning by text decoupling and component alignment. 3D-SPS \cite{luo20223d} regards the visual grounding task as point selecting, points are filtered from coarse to fine by MLP(Multilayer Perceptron) and Transformer layers. Single-stage methods indeed improve the defective outputs from independent detectors, but two-stage methods are still applied widely owing to their rich intermediate results for complex demands in scene understanding tasks. In this paper, we only focus on the matching of the vision and language modalities in the two-stage framework as the same as ReferIt3D \cite{achlioptas2020referit3d}. 

\subsection{Graph Attention for Visual-language}
In order to learn relationship representation to deepen the understanding of complex scene situations or events, graph neural network(GNN) has been used in many cross-modal tasks such as visual question answering(VQA), visual grounding, and image-text match. Graph attention network(GAT) is one of the most popular GNN variants, where every node computes the weight of its neighbors and aggregates relevant features to update its representation \cite{brody2021attentive}. ReGAT \cite{li2019relation} constructs two graphs for detected objects, a full-connected relation graph for implicit relation and a pruned graph with prior knowledge, calculated by attention mechanism. The work in \cite{zhu2021object} for VQA also introduces the graph attention convolution layer, but the graph is based on object difference with extra soft attention. For image-text matching tasks, researchers tend to build the GNNs with image areas and noun phrases respectively to obtain global and local correspondence \cite{long2022gradual,liu2022learning,jing2021learning}. 

GAT is similarly employed in the 3D cross-modal field. FFL-3DOG \cite{feng2021free} generates a context-aware object representation by a 3D visual graph matched with the nodes of the language scene graph, and the top K proposals with higher scores are chosen to predict the grounding target. Nevertheless, in this paper, instead of enhancing the multi-modal fusion on graphs by node match, we add language-guided memory into the graph attention layer to allow the node to learn relevant information automatically.

\begin{figure*}[!t]
	\centering{\includegraphics[width=0.95\linewidth]{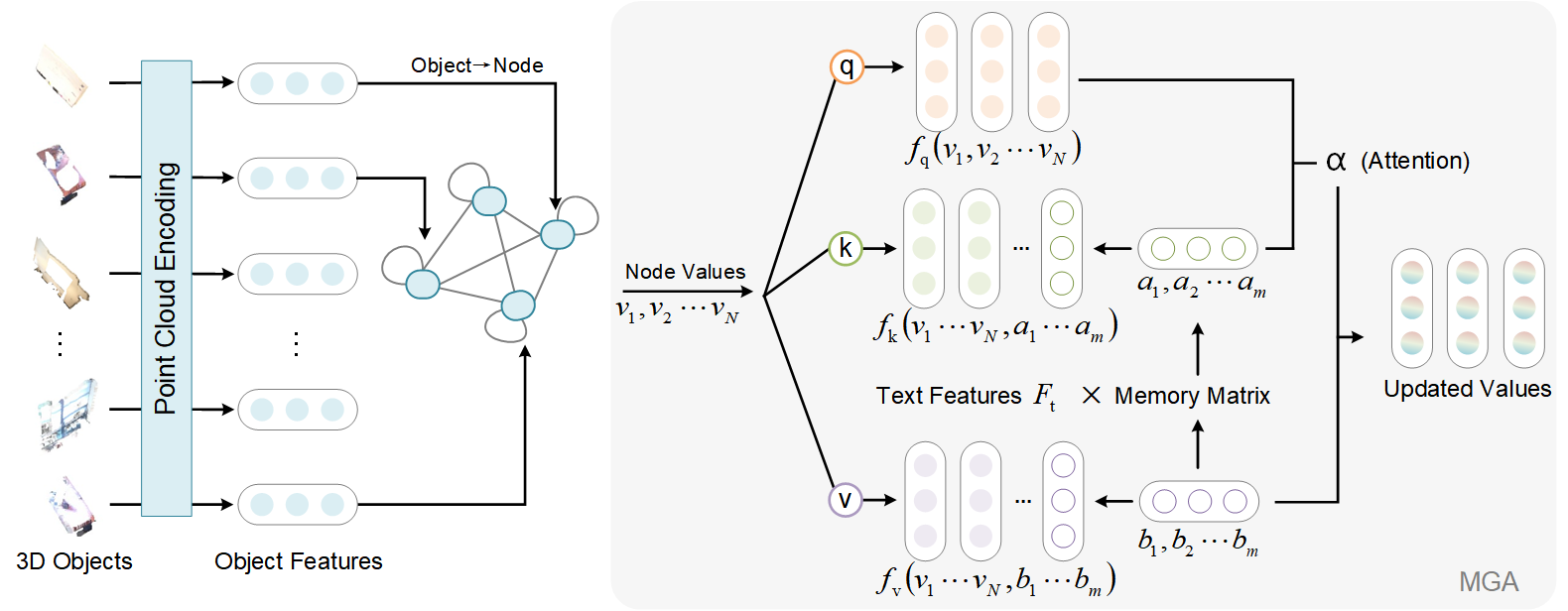}}
	\caption{The implementation process of our proposed memory graph attention(MGA) layer. Given a full-connected graph with object features as nodes, a multi-modal memory unit is added to the key and value of the attention operator to update the node values.}
  \vspace{-0.4cm}
    \label{fig3}
\end{figure*}
\section{Methods}
\subsection{Overview}
Firstly, the 3D scene point clouds are segmented into independent objects for subsequent work. The instance labels in ReferIt3D \cite{achlioptas2020referit3d} are available from the ground truth of ScanNet Dataset \cite{dai2017scannet}. On the contrary, ScanRefer \cite{chen2020scanrefer} needs a prerequisite network to output objects, and we employed pre-trained Pointgroup \cite{jiang2020pointgroup} like previous works \cite{yuan2021instancerefer,huang2022multi} for point cloud segmentation. Fig \ref{fig2} shows the model architecture and inference process of localization. The proposals are further calculated in two important modules, semantic-enhanced visual encoding and relation learning on graph attention. Finally, a Transformer decoder outputs the localization results for the object features from the graph nodes and text features from a pre-trained language model. 

\subsection{Semantic-enhanced Visual Encoding}
For a fair comparison with previous methods like \cite{yang2021sat, roh2022languagerefer, jain2022bottom, he2021transrefer3d, huang2022multi}, we use the same PointNet++ \cite{qi2017pointnet++} as a basic backbone to encode the initial point cloud containing $N$ objects. We randomly sample 1024 points with coordinates and colors $(X, Y, Z, R, G, B)$ for each object and feed them into the network, getting $N$ 768-dimensional features. 

Afterwards, we build a new representation $(X, Y, Z, C)$ for 3D points in the semantic point cloud. It is defined on high-level semantics, where $C$ represents the semantic category. The category information can be obtained from an extra MLP classification module or previous segmentation results. We use a smaller network to encode semantic point clouds without complex color or texture. It is guided to understand more relationships by simplified representation of objects, rather than limited to appearance learning. Subsequently, the encoding features $V_{rgb}$ and $V_{sem}$ from the two point clouds are fused as follows:
\begin{align}
V_F&=Relu([fc_1 (V_{rgb}),fc_2 (V_{sem})]) \label{eq1}
\end{align}
where $[,]$ represents the concatenation, $fc_1$ and $fc_2$ are the full-connected layers for feature mapping, and their results are concatenated to be the semantic-enhanced feature $V_F$.

In this cascade structure, the generation of a semantic point cloud is completely determined by the previous category information. For visual grounding tasks where segmentation categories cannot be employed directly (like Referit3D), the classification results in the early training stage are inaccurate, which causes the semantic encoder to match information in turbid modes and affects the optimization direction of subsequent modules. Consequently, the shallow encoder for RGB point clouds is pre-trained with the same datasets before the holistic training to overcome this defect. 

\subsection{Relation Graph Learning}
The core of understanding referential descriptions is to grasp the position relationships of the mentioned objects. We construct a full-connected graph to autonomously learn implicit relationships among objects based on a graph attention network. The object features extracted in the previous networks constitute the nodes of the scene graph. In our designed graph network, each node undergoes two multi-head attention layers. The $i$-th node value is updated to $v_i^{'}$ by information aggregation of adjacent nodes $D_i$:
\begin{align}
v_i^{'}&=\rVert_{k=1}^{K}\sigma (\sum_{j\in D_i} \alpha_{ij}^{k} W^{k} v_j) \label{eq2}
\end{align}
where $v_j$ represents the node value of $j$-th neighbor, $\alpha_{ij}^{k}$ is the attentive weight of $k$-th head, $W^{k}$ is the projection matrix, and $\sigma()$ indicates a nonlinear activation layer. The results of K heads are weighted and added to obtain the final value. To enable the relational learning model to leverage textual information and better adapt to view transformations, we propose two sub-modules to improve the intrinsic attention algorithm. The node update process in one graph attention layer is shown in Fig \ref{fig3}. 
 
\subsubsection{Auxiliary Memory Unit}
Inspired by the consideration of attention limitation in \cite{cornia2020meshed}, the self-attention coefficient $\alpha$ in equation \eqref{eq2} is weak to inherit the prior knowledge about real scenes. We add a learnable matrix as a memory unit into the key and value of the attention operator to avoid this limitation. However, this attention structure is still a probabilistic model purely relying on vision. For targets grounded by multiple reference relationships or composite relationships, it is necessary to introduce text modality in relational learning to select attention content and reduce redundant connections. In other words, graph nodes will perform multi-directional differentiated learning on the description-related information flow during the message transmission and updating, not only from the target to the referred objects. Therefore, we design a novel memory unit that aligns text information to improve graph attention:
\begin{align}\label{eq3}
X_m &=[[v_{i\in N}],F_t\times M]
\end{align}
where $[v_{i\in N}]$ is the feature matrix composed of $N$ object features, the text feature vector $F_t$ and a learnable memory matrix $M$ are fused as a text-enhanced memory unit and combined with original features. The new feature matrix $X_m$ is immediately used for graph attention computation to obtain the updated node values $v_{i\in N}^{'}$: 
\begin{align}\label{eq4}
[{v_{i\in N}^{'}}] &=softmax(\frac{W_q [v_{i\in N}] \cdot W_k X_{m}^{T}}{\sqrt{d_m}} ) W_v X_{m}
\end{align}
where $W_q$, $W_k$ and $W_v$ are weight coefficients, the dimension of $X_m$ is used as a scaling factor $d_m$.  

\subsubsection{Multi-view Position Embedding}
The description of a 3D scene may depend on observations from any perspective, so deep features extracted from point clouds need to adapt different views. In terms of each object, geometry, texture, or color attributes in point-form expression are inherently not affected by the view. Due to the impact of object location on relational learning with referential languages, we encode multi-view information into position embedding of graph attention. Specifically, multiple object coordinates are generated under the isometrically rotated scene views, and the point cloud features are decoupled from them as shared information. We encode the pairwise position relationship and add it to the calculation of the attention coefficient.
\begin{align}\label{eq5}
E_r &=[sin(\frac{M_r}{\mu_a}),cos(\frac{M_r}{\mu_a})]\\
\alpha_r^{'} &= max\{E_r,0\} \alpha_r
\end{align}
where $M_r$ express the relative relationship of object$<i,j>$ from the perspective $r$, calculated as ($\log \frac{x_i-x_j}{l_i^{X}}$, $\log \frac{y_i-y_j}{l_i^{Y}}$, $\log \frac{z_i-z_j}{l_i^{Z}}$), $l_i$ is the length of the bounding box of the $i$-th object on each axis. $M_r$ is calculated by sin and cos operators at wavelength $\mu_a$ to obtain the position encoding vector $E_r$, which is superimposed on the attention coefficient $\alpha_r$ as a position-related feature.  

\begin{figure}[!t]
	\centering
    \includegraphics[width=0.7\linewidth]{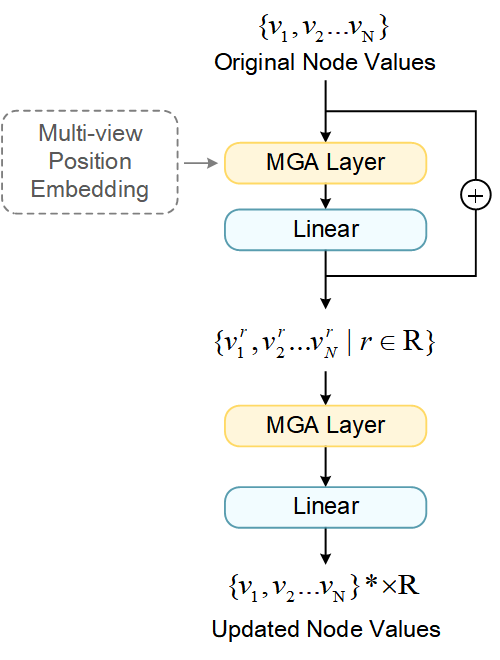}
	\caption{The illustration of a two-layer graph network structure based on MGA layers. Positioning embedding of $R$ views only works in the first layer.}
    \label{fig4}
\end{figure}

As shown in Fig \ref{fig4}, our proposed graph network consists of memory graph attention(MGA) layers and linear layers. Geometric features from $R$ views are embedded into the first-layer attention operator, and each node produces specific values under the guidance of $r$-th view. The original node value is added to the updated value of the first layer as input to the next layer. Finally, the output features are aggregated by the average operator to obtain visual encoding features containing semantic and relational information.

\begin{figure*}[!t]
	\centering{\includegraphics[width=1\linewidth]{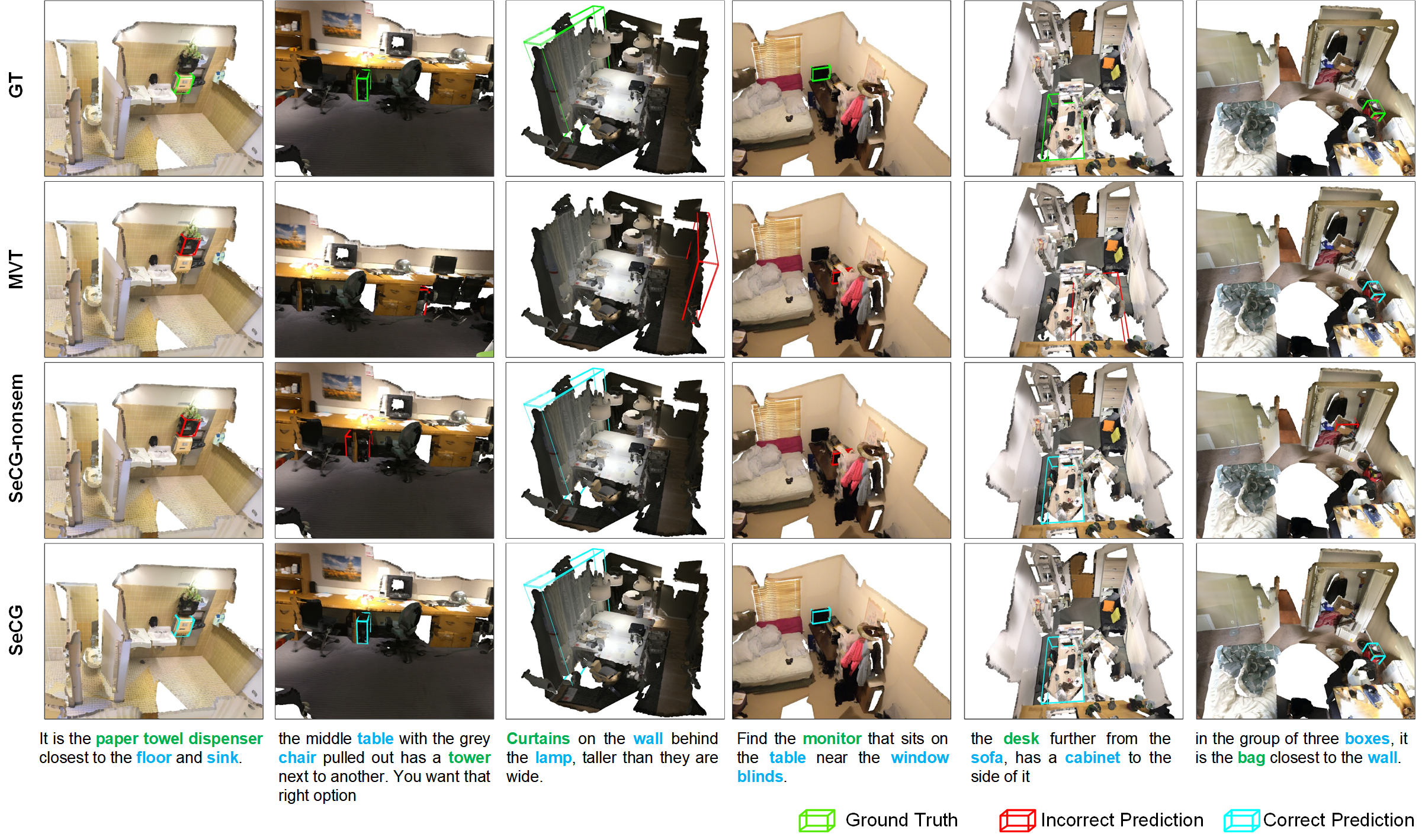}}
	\caption{Visualization results on visual grounding samples with multiple referred objects from Nr3D. The words represent targets are highlighted in green and the reference words are highlighted in blue. Our localization results are compared with MVT \cite{huang2022multi} that directly matches text features with multi-view visual features.}
    \label{fig5}
\end{figure*}

\subsection{Visual-Language Training}
The key to localization is the matching of visual content and text description. We have obtained the object features from point clouds as visual encoding results. A pre-trained language model with BERT \cite{kenton2019bert} is employed for language encoding, fine-tuned with a lower learning rate in the training stage. The sentences in datasets are directly tokenized and encoded to generate 768-dimensional text features. In the localization stage, we use a standard Transformer decoder consisting of multi-head attention layers to match objects and descriptions. The expression of the decoding layer is as follows:
\begin{align}\label{eq6}
D(F_v,F_l) &= \rVert_{i=1}^{M} A_n(W_i^{q} A_s(F_v),W_i^{k} F_l,W_i^{v} F_l)
\end{align}
where $F_v$ and $F_l$ are visual coding and language coding features respectively, $A_s$ represents the self-attention calculation of $F_v$, $W_i^{q}$, $W_i^{k}$ and $W_i^{v}$ represent the three linear transformations on query, key and value of the attention operator $A_n$ in the $i$-th head. After stacking several decoding layers, the probability of each object matching the described target is exported. In addition to computing the localization loss of the target, we also add classification layers regarding the object types and description sentences. The loss is defined as,
\begin{align}\label{eq7}
Loss &= \lambda_1 l_{sem}+\lambda_2 l_{lan}+\lambda_3 l_{ref}
\end{align}
where $l_{ref}$ is the location loss, $l_{sem}$ is the classification loss of objects from semantic point cloud, $l_{lan}$ is the loss about classifying the described target through sentence encoding features, $\lambda_1-\lambda_3$ are weight coefficients. All of them are computed using the cross-entropy loss function on linearly mapped features.

\begin{table*}[!t]
\begin{center}
	\caption{Comparison with other recent methods on Nr3D and Sr3D}
    \renewcommand\arraystretch{1.3}
    \tabcolsep=8pt
    \label{tab1}
    \begin{threeparttable}
	\begin{tabular}{c|ccccc|ccccc}
    \Xhline{2pt}
	  \multirow{2}{*}{Methods}&\multicolumn{5}{c|}{\textbf{Nr3D}} &\multicolumn{5}{c}{\textbf{Sr3D}}\\
        \cline{2-11}
		&Overall   &Easy    &Hard  &V-dep  &V-indep &Overall   &Easy    &Hard  &V-dep  &V-indep \\
       \cline{1-11}
		ReferIt3D \cite{achlioptas2020referit3d}  &35.6\% &43.6\%  &27.9\% &32.5\% &37.1\% &40.8\% &44.7\% &31.5\% & 39.2\% &40.8\% \\
	  TGNN \cite{huang2021text}   &37.3\% &44.2\%  &30.6\% &35.8\% &38.0\% &45.0\% &48.5\% &36.9\% & 45.8\% &45.0\% \\
		InstanceRefer \cite{yuan2021instancerefer} &38.8\% &46.0\% &31.8\% &34.5\% &41.9\% &48.0\% &51.1\% &40.5\% &45.4\% &48.1\% \\
		3DVG-Trans \cite{zhao20213dvg}  &40.8\%  &48.5\% &34.8\% &34.8\% &43.7\% &51.4\% &54.2\% &44.9\% &44.6\% &51.7\% \\
        TransRefer3D \cite{he2021transrefer3d} &42.1\%  &48.5\% &36.0\% &36.5\% &44.9\% &57.4\% &60.5\% &50.2\% &49.9\% &57.7\% \\
        LanguageRefer \cite{roh2022languagerefer}   &43.9\%  &51.0\% &36.6\% &41.7\% &45.0\% &56.0\% &58.9\% &49.3\% &49.2\% &56.3\% \\
        SAT \cite{yang2021sat}   &49.2\%  &56.3\% &42.4\% &46.9\% &50.4\% &57.9\% &61.2\% &50.0\% &49.2\% &58.3\% \\
        BUTD-DETR \cite{jain2022bottom}   &54.6\%  &60.7\% &48.4\% &46.0\% &58.0\% &67.0\% &68.6\% &\textbf{63.2\%} &53.0\% &67.6\% \\
        MVT \cite{huang2022multi}  &55.1\%  &61.3\% &49.1\% &54.3\% &55.4\% &64.5\% &66.9\% &58.8\% &\textbf{58.4\%} &64.7\% \\
        ViewRefer \cite{guo2023viewrefer}  &56.0\%  &63.0\% &49.7\% &55.1\% &56.8\% &67.0\% &68.9\% &62.1\% &52.2\% &67.7\% \\
        \hline
        SeCG(ours)   &\underline{\textbf{57.9\%}}  &\textbf{64.2\%} &\textbf{51.9\%} &\textbf{57.2\%} &\textbf{58.3\%} &\underline{\textbf{68.3\%}} &\textbf{71.1\%} &61.7\% &57.0\% &\textbf{68.8\%} \\
    \Xhline{2pt}
    
	\end{tabular}
    \begin{tablenotes}   
     \item The first-ranked overall evaluation results are underlined, and the highest item for each indicator is bolded.
    \end{tablenotes}
 \end{threeparttable}
\end{center}
\end{table*}

\begin{table*}[!t]
\begin{center}
    \caption{Comparison with other recent methods without extra data on Scanrefer.}
    \renewcommand\arraystretch{1.3}
    \tabcolsep=8pt
    \label{tab2}
    \begin{threeparttable}
	\begin{tabular}{c|c|c|cccccc}
    \Xhline{2pt}
	  \multirow{2}{*}{Methods}&\multirow{2}{*}{Detector}&\multirow{2}{*}{2D}&\multicolumn{2}{c}{Unique}&\multicolumn{2}{c}{Multiple}&\multicolumn{2}{c}{Overall}\\
        \cline{4-9}
		&   & &Acc@0.25&Acc@0.5&Acc@0.25&Acc@0.5&Acc@0.25&Acc@0.5\\
       \hline
       \multicolumn{8}{c}{Validation Results}\\
       \hline
	ScanRefer \cite{chen2020scanrefer} &VoteNet &&67.64\% &46.19\%  &32.06\% &21.26\% &38.97\% &26.10\%\\
        TGNN \cite{huang2021text} &3D-UNet & &68.61\% &56.80\%  &29.84\% &23.18\% &37.37\% &29.70\% \\
        SAT \cite{yang2021sat} &VoteNet &\checkmark &73.21\% &50.83\%  &37.64\% &25.16\% &44.54\% &30.14\% \\
	InstanceRefer \cite{yuan2021instancerefer} &PointGroup & &77.45\% &66.83\% &31.27\% &24.77\% &40.23\% &32.93\%\\
        MVT \cite{huang2022multi} &PointGroup &&77.67\%  &66.45\% &31.92\% &25.26\% &40.80\% &33.26\% \\
        ViewRefer \cite{guo2023viewrefer} &PointGroup &&-  &- &33.08\% &26.50\% &41.30\% &33.66\%\\
	3DVG-Transformer \cite{zhao20213dvg} &VoteNet &&77.16\%  &58.47\% &38.38\% &28.70\% &45.90\% &34.47\%\\
        3DJCG* \cite{cai20223djcg} &VoteNet &\checkmark &-  &64.50\% &- &30.29\% &- &36.93\%\\
        \hline
        SeCG(ours)   &PointGroup &&\textbf{77.72\%} &\textbf{67.41\%} &\textbf{38.53\%} &\textbf{31.67\%} &\textbf{46.13\%} &\underline{\textbf{38.59\%}}\\
        \hline
        \multicolumn{8}{c}{Test Results}\\
        \hline
        ScanRefer \cite{chen2020scanrefer} &VoteNet &\checkmark &68.59\% &43.53\%  &34.88\% &20.97\% &42.44\% &26.03\%\\
        TGNN \cite{huang2021text} &3D-UNet & &68.34\% &58.94\%  &33.12\% &25.26\% &41.02\% &32.81\% \\
        InstanceRefer \cite{yuan2021instancerefer} &PointGroup & &77.82\% &\textbf{66.69\%} &34.57\% &26.88\% &44.27\% &35.80\%\\
        D3Net* \cite{zhenyu2021d3net} &PointGroup &\checkmark &76.59\%  &65.79\% &36.19\%&27.26\%&45.25\%&35.90\%\\
        \hline
        SeCG(Ours)   &PointGroup & &72.88\% &61.75\% &\textbf{36.96\%} &\textbf{29.33}\%&45.01\% &\underline{36.60\%}\\
        SeCG+(Ours)   &PointGroup &\checkmark &\textbf{77.99}\% &66.28\% &36.36\% &28.23\%&\textbf{45.69\%}&\underline{\textbf{36.77\%}}\\
    \Xhline{2pt}
	\end{tabular}
    \begin{tablenotes}   
    \item The first-ranked overall evaluation results are underlined, and the highest item for each indicator is bolded.
    \end{tablenotes}
 \end{threeparttable}
\end{center}
\end{table*}

\section{Experiment}
\subsection{Datasets}
Nr3D and Sr3D are two large-scale and complementary visual grounding datasets from ReferIt3D \cite{achlioptas2020referit3d}. Nr3D contains 41,503 natural utterances about object description generated by taggers on ScanNet \cite{dai2017scannet} 3D scenes. Sr3D is a synthetic dataset where referential languages are automatically generated by templates of five established spatial relationships, containing 83,572 utterances. The above-described objects have no more than six distractors of the same type in the corresponding scene. 

ScanRefer dataset \cite{chen2020scanrefer} provides 51,583 human-written free-form descriptions of 11,046 objects in ScanNet. Compared with ReferIt3D, it prefers to use multiple short sentences to describe one object and usually requires the introduction of an object detector to generate proposals before localization due to the unknown object bounding boxes. 

\subsection{Evaluation Metrics}
In the visual grounding task, each description corresponds to a unique target, the number of correctly matched text-object pairs determines the accuracy of the localization models. As the datasets give different challenges, the evaluation metrics are also diverse. 

Nr3D is divided into easy and hard subsets based on whether there are more than 2 distractors to evaluate the model's fine-grained discrimination ability of similar objects. On the other hand, some referential languages are defined according to a specific perspective, so Nr3D also evaluates the robustness of perspective changes from view-dependent and view-independent samples respectively. In addition, the major problem discussed in this paper is multi-relationship utterances with different references. We employ the number of non-target type objects mentioned in the sentence to evaluate it.

In ScanRefer, the object bounding boxes are also included as prediction items in the evaluation criteria. When calculating the localization accuracy, it is necessary to judge whether the prediction box correctly hits the target based on IoU with thresholds of 0.5 and 0.25. ScanRefer sets two subsets, unique and multiple samples, to evaluate model performance, where the unique means only a single object of the target class in the scene. 

\subsection{Implementation Details}
Both point cloud encoders are based on PointNet++ and the semantic point cloud encoder is a smaller one than the common structure. The relation graph network consists of two MGA layers, where the feature dimension of each node is set to 768, the number of heads $K$ is 8, the height of memory matrix $M$ is 10, and the position embedding applies 4 views. The loss weight coefficients $\lambda_1$ and $\lambda_2$ are set to 0.5, $\lambda_3$ is set to 1.0. All models are trained with a batch size of 36 for 120 epochs, using the Adam \cite{kingma2014adam} optimizer. The leaning rate is initialized to $5 \times 10^{-4}$ with decay of 0.65
every 10 epochs from 30 to 80 epochs, and the pre-trained classification layers, language encoding layers, and transformer layers are $1/10$ of that. 

\begin{figure*}[!t]
	\centering{\includegraphics[width=1\linewidth]{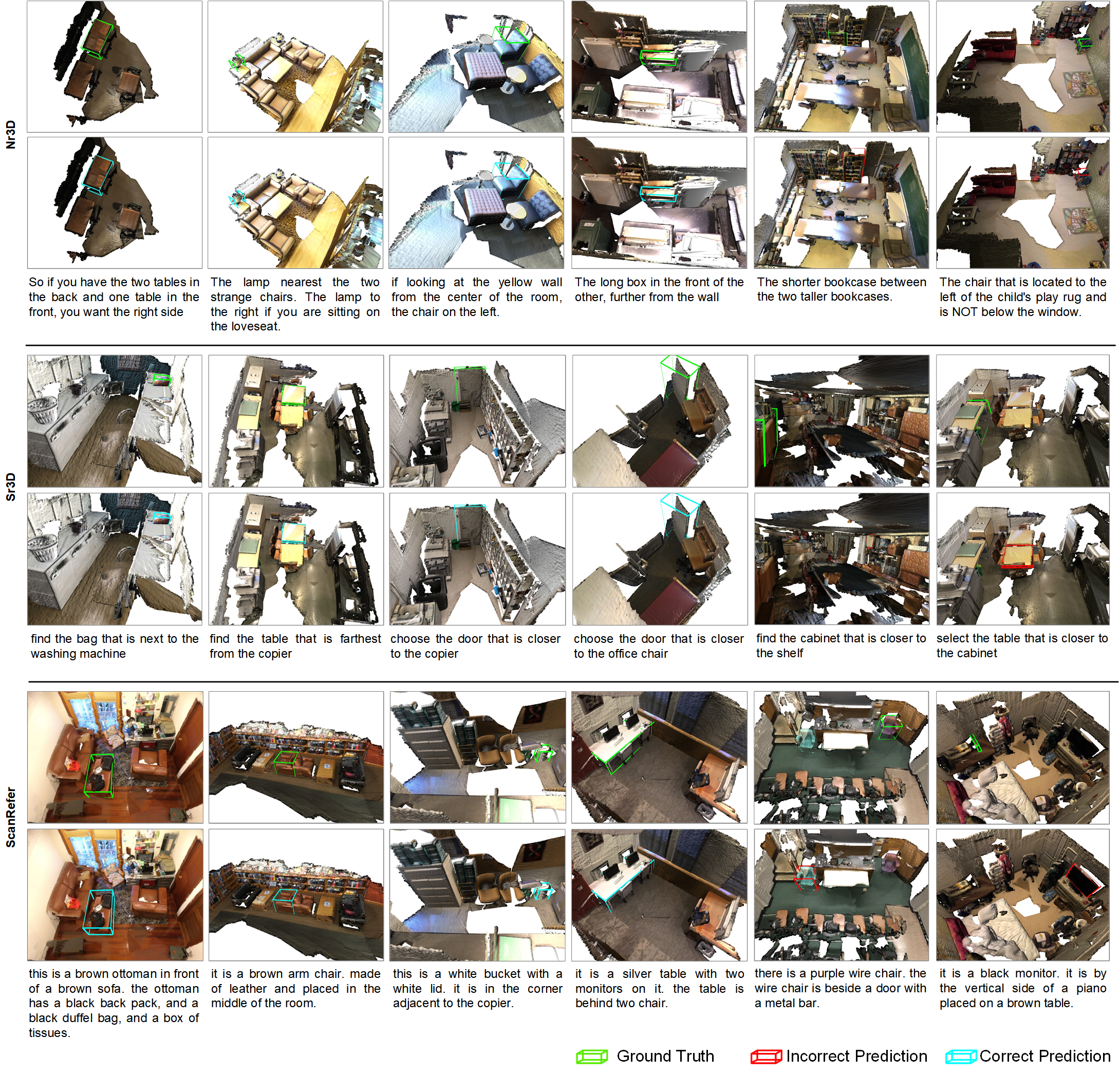}}
	\caption{Visualization results of our method on three datasets. In all paired examples, the first row is the ground truth and the second is the predicted result. In Nr3D and Sr3D the shape of the predicted box is exactly the same as the ground truth box. In ScanRefer the predicted box and the ground truth box are considered to match when their IoU is greater than 0.5.}
    \label{fig6}
\end{figure*}

\subsection{Localization Results}
\subsubsection{Nr3D/Sr3D}
Table \ref{tab1} shows the performance of our method and recent works on Nr3d and Sr3d. \enquote{V-dep} and \enquote{V-indep} represent the evaluation results of view-dependent and view-independent samples. \enquote{Easy} indicates the localization accuracy on samples of less than 2 objects with the same category of target in the same scene, \enquote{Hard} is the opposite. Our proposed SeCG reaches the state-of-the-art with overall accuracy of 57.9\% and 68.3\% on both datasets, especially improving on hard and view-dependent samples of Nr3D with 2.2\% and 2.1\%. Compared with synthetic sentences in Sr3D generated with pairing-relation templates, the human-defined referential relations in Nr3D are more complex and even require indirect relation to complete the localization. Therefore, our main improvement in relational learning is most evident in Nr3D. This shows that our approach effectively analyzes and resolves issues involving multiple reference relationships. 

Fig \ref{fig5} shows some visualization results on samples with more than 2 references. Our approach adds a relational learning module and semantic enhancement module based on MVT, using a similar backbone and multi-view aggregation technique. It's noticed that MVT uses language-guided supervision in training with 525 instance classes to improve the overall accuracy, but we only set 40 rough classes in the loss. Because we hope the model focuses more on class-independent understanding. \enquote{SeCG-nonsem} represents our model trained without semantic enhancement. Our proposed graph attention network is a definite structure for understanding referential relationships, and semantic-level encoding is also conducive to the extraction of relational information. 

\subsubsection{ScanRefer}
Our method is compared with other two-stage visual grounding methods on ScanRefer in Table \ref{tab2}. \enquote{Acc@0.25} and \enquote{Acc@0.5} represent the accuracy of the predicted bounding boxes with 0.25 and 0.5 IoU thresholds respectively. \enquote{\checkmark} marks methods that incorporate multi-view 2D features during inference, and unremarkable methods only use 3D point clouds. \enquote{*} means that the model is only trained on visual grounding datasets and losses in a joint architecture. \enquote{Test Results} only includes published methods that have been evaluated on the online benchmark. The second column shows the detectors of the first stage and the third column shows whether the multi-view features from 2D images are added. Our proposed method outperforms others in overall accuracy on both the validation and test sets. For the models that also use PointGroup as the detector, we have significantly improved on the hard samples with multiple distractors. It is noticed that InstanceRefer has a particularly high accuracy on \enquote{Unique} with 66.69\% mainly due to its filter of objects by target category before localization, which greatly reduces the difficulty of single-object scenes. 

In the online benchmark, we test our methods with and without 2D information. SeCG is to use point clouds only, and SeCG+ is to project 2D multi-view features of pre-trained ENet\cite{paszke2016enet} onto points aligned to ScanRefer baseline. Table \ref{tab2} indicates that adding 2D features greatly helps localizing the unique object, but is not conducive to scenarios with multiple same-class objects. That is because the latter needs more relationships to support its description while redundant appearance features may interfere with the relational learning direction, which is consistent with our previous analysis of semantic effect. 

Some localization results are visualized in Fig \ref{fig6}. Based on the results of various datasets, our method can correctly locate in most complex scenarios but has a weak understanding of some rare attributes and relationships. For example, ”shorter” implicitly compares the target to other similar objects but is not mentioned directly, and some negative descriptions with ”not” may cause misleading. In addition, the incorrect predictions usually locate similar objects near the target. Improvements can be made in these directions in the future.

\begin{table}[!t]
\begin{center}
	\caption{Ablation studies of the graph attention module}
    \renewcommand\arraystretch{1.3}
    \label{tab3}
	\begin{tabular}{ccc|ccc}
    \Xhline{2pt}
        GAT & MP & MU & Overall &M-Num $\textgreater$ 2 &M-Num $\leq$2 \\
        \cline{1-6}
            &            &           & 51.8\%  &45.6\%  & 52.2\%  \\
        \checkmark    &            &           & 53.9\%  & 52.4\%  & 54.3\%  \\
        \checkmark    & \checkmark &           & 54.6\%  & 53.4\%  & 54.9\%  \\
         \checkmark    & \checkmark & \checkmark & 55.2\%  & 54.3\%  & 55.5\%  \\
    \Xhline{2pt}
	\end{tabular}
\end{center}
\end{table}

\subsection{Ablation Study}
\subsubsection{Relation Graph}
We verify the effectiveness of each sub-module in the relation graph network with ablation experiments in Table \ref{tab3}. \enquote{GAT}, \enquote{MU}, and \enquote{MP} represent the graph attention module, memory unit module, and multi-view position embedding module. \enquote{M-Num} is the number of mentioned objects under rough statistics by parsing the object classes that occurred in utterances. The evaluation is based on Nr3D, whose unified object boxes and free-form utterances exclude the influence of the detector and directly reflect the ability to understand real languages. Compared to the baseline in row 1, our proposed graph network has improved the grounding performance on samples with more than 2 object classes mentioned in the description for 8.7\%, outstripping other single-relation samples. In other words, when there are at least two direct or indirect referential relations in the utterance, our model can better understand them and locate the target.

\subsubsection{Semantic Enhancement}
Table \ref{tab4} separately lists the influence of semantic-enhanced encoding mode with different backbones. PoinetNet++ is the most used network in 3D visual grounding tasks. For a fair comparison with other methods, we use it to evaluate the experiment results. Point Transformer \cite{zhao2021point} is a more novel point cloud network that uses the attention mechanism. Smaller networks (0.91M of PointNet++ and 1.53M of Point Transformer) are constructed to capture more location information for relational learning from semantically rendered point clouds. It can be seen that semantic point cloud encoding has improved the localization performance of both backbones, especially on hard samples. Point Transformer has a larger scale but no obvious advantages, which indicates that relational learning requires directional information such as semantic-level position extraction more than rich features.

\begin{table}[!t]
\begin{center}
	\caption{Ablation studies of semantic-enhanced encoding module}
    \renewcommand\arraystretch{1.3}
    \label{tab4}
	\begin{tabular}{c|c|ccc}
    \Xhline{2pt}
        Backbone & Sem-encoder & Overall & Easy  & Hard \\
        \cline{1-5}
        PointNet++ &  & 55.2\%  &  61.6\% & 49.1\% \\
        PointNet++ & \checkmark & 57.9\%  &  64.2\%  & 51.9\% \\
        \cline{1-5}
        Point Transformer&  & 56.4\%  &  62.3\%  &  50.3\%\\
        Point Transformer& \checkmark  & 57.2\% & 63.4\%  & 51.2\% \\
    \Xhline{2pt}
	\end{tabular}
\end{center}
\end{table}

\section{Conclusion}
In this paper, we point out the challenge of 3D visual grounding about weak understanding of multiple referred objects. To perceive complex and indirect relationships of the mentioned objects, SeCG, a semantic-enhanced visual grounding model based on graph attention is proposed. We construct a cross-modal graph attention network with a language-guided updating layer for relational learning and utilize prior semantic knowledge to enhance its perception. Different from previous works that directly match visual and language features, our proposed 3D encoding module can provide more useful relationship information and improve the matching in complex referential descriptions. Experimental results on ReferIt3D and ScanRefer show that our method has outperformed others in overall accuracy, especially on the targets that require multiple references to locate. In the current performance, unintuitive or uncommon relationship descriptions are still challenging to comprehend. We will improve our model to improve text understanding and cross-modal alignment in the future.


\bibliographystyle{IEEEtran}
\bibliography{literature}

\end{document}